\documentclass[10pt,twocolumn,letterpaper]{article}

\usepackage[pagenumbers]{cvpr} % To force page numbers, e.g. for an arXiv version

\usepackage{graphicx}
\usepackage{amsmath}
\usepackage{amssymb}
\usepackage{booktabs}

\usepackage[pagebackref,breaklinks,colorlinks]{hyperref}
\usepackage{algorithm}
\usepackage{algorithmic}
\usepackage{epsfig}
\usepackage{amsmath}
\usepackage{amssymb}
\usepackage{booktabs}
\usepackage{multirow}
\usepackage{booktabs}
\usepackage[table,xcdraw]{xcolor}
\usepackage{soul}
\makeatletter
\@namedef{ver@everyshi.sty}{}
\makeatother
\usepackage{tikz}
\usepackage{pifont}
\newcommand{\cmark}{\ding{51}}%
\newcommand{\xmark}{\ding{55}}%

\newcommand{\modelname}{Gaussian Approximated Post-processing}
\newcommand{\shortname}{GAP}

\usepackage[capitalize]{cleveref}
\crefname{section}{Sec.}{Secs.}
\Crefname{section}{Section}{Sections}
\Crefname{table}{Table}{Tables}
\crefname{table}{Tab.}{Tabs.}

\def\cvprPaperID{9954} % *** Enter the CVPR Paper ID here
\def\confName{CVPR}
\def\confYear{2023}

\begin{document}

\title{Post-Processing Temporal Action Detection}

% \and
\author{Sauradip Nag$^{1,2}$
\and
Xiatian Zhu$^{1,3}$
\and
Yi-Zhe Song$^{1,2}$
\and 
Tao Xiang$^{1,2}$ 
\and \newline
{\small $^1$ CVSSP, University of Surrey, UK} ~ 
{\small $^2$ iFlyTek-Surrey Joint Research Center on Artificial Intelligence, UK} \\
{\small $^3$ Surrey Institute for People-Centred Artificial Intelligence, UK} ~
}
\maketitle

%%%%%%%%% ABSTRACT
\begin{abstract}
   Existing Temporal Action Detection (TAD)  methods typically take a pre-processing step in converting an input varying-length video into a fixed-length snippet representation sequence, before temporal boundary estimation and action classification.
This pre-processing step would {\em temporally downsample} the video, reducing the inference resolution and hampering the detection performance in the original temporal resolution.
In essence, this is due to a temporal quantization error introduced
during the resolution downsampling and recovery.
This could negatively impact the TAD performance, but  is largely ignored by existing methods.
To address this problem,
in this work we introduce a novel {\em model-agnostic} post-processing method 
% for alleviating this low-resolution inference issue of existing TAD methods 
without model redesign and retraining.
Specifically, we model the start and end points of action instances with a Gaussian distribution for enabling %
temporal boundary inference at a {\em sub-snippet} level. %
% By treating the model inference as a black-box procedure,
% our method can be applied generally to any existing pre-trained TAD methods,
% serving as a generic post-processing plug-in.
%
We further introduce an efficient Taylor-expansion based approximation, dubbed as {\em Gaussian Approximated Post-processing} ({\bf GAP}).
Extensive experiments demonstrate that our GAP can consistently improve a wide variety of pre-trained off-the-shelf TAD models on the challenging ActivityNet (+0.2\%$\sim$0.7\% in average mAP) and THUMOS (+0.2\%$\sim$0.5\% in average mAP) benchmarks. \textcolor{black}{
Such performance gains are already significant
and highly comparable to those achieved by novel model designs.}
% Since TAD is a saturating field and also uses a very challenging metric (average mAP over IoU thresholds from 0.5 to 0.95 for ActivityNet and from 0.3 to 0.7 for THUMOS), the improvement from \shortname{} is significant in the community and shows similar/better improvement than major recent state-of-the-art methods.
% , without any algorithmic modification and model retraining.
Also, %through extensive experiments we observed that 
GAP can be integrated with model training 
for further performance gain.
% improving the overall performance by a significant margin. 
%
Importantly, GAP enables lower temporal resolutions
for more efficient inference,
facilitating low-resource applications. The code will be available in \url{https://github.com/sauradip/GAP}
\end{abstract}

%%%%%%%%% BODY TEXT

\section{Introduction}
The objective of Temporal action detection (TAD) is to identify both the temporal interval (\ie, start and end points) and the class label of all action instances in an untrimmed video \cite{idrees2017thumos,caba2015activitynet}. 
Given a test video, existing TAD methods typically  generate a set of action instance candidates via {proposal generation} based on
regressing predefined anchor boxes 
\cite{xu2017r,chao2018rethinking,gao2017turn,long2019gaussian}
or 
directly predicting the start and end times of proposals
\cite{lin2019bmn,buch2017sst,lin2018bsn,xu2020g,nag2021few,xu2021boundary,xu2021low} and global segmentation masking \cite{nag2022gsm}. 
\begin{figure}[t]
    \centering
    \includegraphics[scale=0.6]{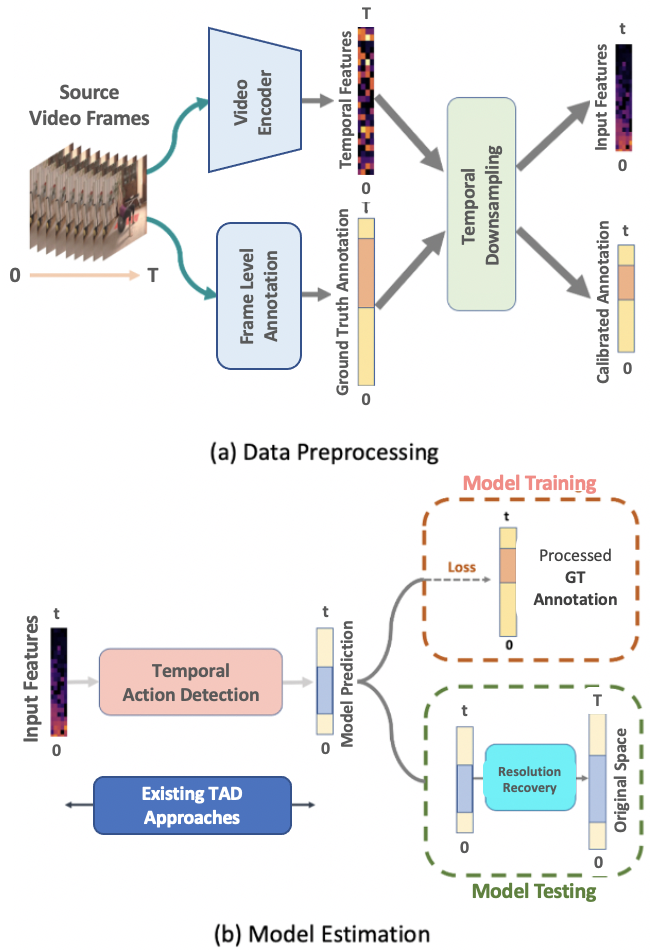}
    % \vspace{0.2in}
    \caption{A typical pipeline for temporal action detection. 
    (a) For efficiency and model design ease, temporal resolution reduction is often applied % in the frame-level feature space 
    during pre-processing.
    This causes model inference at lower (coarse) temporal resolutions.
    (b) After bringing the prediction results back to the original temporal resolution during inference, quantization error will be introduced inevitably.
    % Whilst hampering the model performance clearly, this problem is totally ignored in the literature.
    % on the original frame-level features and also the action boundary ground-truth annotations. So, the model operates in a low-resolution temporal space which reduces model inference cost significantly. At test time, a temporal resolution recovery is required for the predicted output to map it back to the original video space. 
    }
    \label{fig:1}
    \vspace{-0.7cm}
\end{figure}
To facilitate deep model design and improve computational efficiency,
all TAD methods would pre-process a varying-length video 
into a fixed-length snippet sequence by first extracting frame-level visual features with a frozen video encoders %(\eg, I3D \cite{carreira2017quo} and  TSN \cite{wang2016temporal})
and subsequently sampling a smaller number of feature points (\ie, snippet) evenly (see Fig.~\ref{fig:1}(a)).
% The de-facto standard to solve this problem is first the raw videos are passed into a frozen video encoders (\eg, I3D\cite{} and  TSN \cite{}) and the spatio-temporal features are extracted for the entire video. With the feature label representation, one major obstacle is that, the computational cost is a quadratic function
% of the temporal resolution, preventing the CNN models
% from processing the typically high-resolution raw video
% data. To be computationally affordable, a standard strategy
% (see Fig.~\ref{fig:1}(a)) is to down-sample all the arbritarily large 1-D temporal features into a prefixed small
% resolution with a data preprocessing procedure, before being fed into a temporal action detection model. 
As a result, a TAD model performs the inference 
at {\em lower temporal resolutions}.
% as compared to the original one.
%
%
This introduces a {\bf\em temporal quantization error} that
could hamper the model performance.
For instance, when decreasing
video temporal resolution from 400 to 25, the performance of BMN \cite{lin2019bmn} degrades significantly from 34.0\% to 28.1\% in mAP on ActivityNet.
Despite the obvious connection between the error and performance degradation, this problem is largely ignored by existing methods.
% never been systematically investigated to our best knowledge.

% predicts a start/end point for the actions during inference in the reduced temporal space. This start/end points are then multiplied with the video-duration as a part of resolution recovery process to obtain the start/end points of action proposals/mask in the original video space. 
% For instance,
% with the state-of-the-art model BMN \cite{lin2019bmn}, the aforementioned shifting operation of coordinate encoding brings gain as high as xx\%/xx\% in avg mAP on the challenging ActivityNet and THUMOS dataset (Table 1). It is noteworthy to mention that, this gain is already much more significant than those by most individual art methods.
% But it is never well noticed and carefully investigated in the literature to our best knowledge.

% Contrary to the existing temporal action detection studies,
% in this work we dedicatedly investigate the problem of action boundary representation including encoding and decoding.
% Moreover, we recognise that the temporal resolution is one
% major obstacle that prevents the use of smaller input resolution for faster model inference. When decreasing the
% input temporal resolution from 400-dim to 25-dim, the model performance of BMN \cite{lin2019bmn} drops significantly from XX\% to XX\% on the ActivityNet dataset, although the model
% inference cost falls from xx
% to xx FLOPs.

\begin{figure}
    \centering
    \includegraphics[scale=0.39]{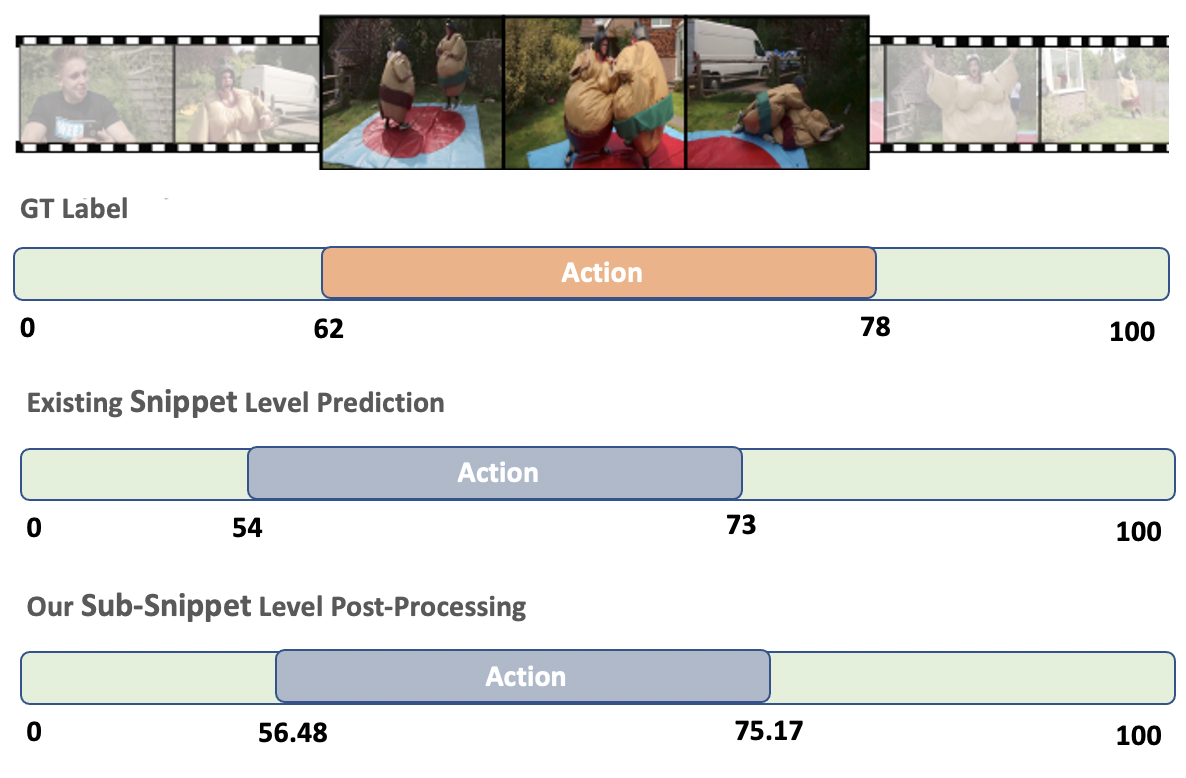}
    \caption{
    {Conventional {\em snippet-level} TAD inference along with our proposed {\em sub-snippet-level} post-processing}. 
    % Existing TAD approaches predict at the snippet-level, our GAP process such action proposals/masks at the sub-snippet level, thus more fine-level foreground prediction.
    }
    \vspace{-0.2in}
    \label{fig:subsnip}
\end{figure}
In this work, we investigate the {\em temporal quantization error} problem 
% with generic TAD methods 
from a post-processing perspective.
Specifically, we introduce a model-agnostic post-processing approach
for improving the detection performance of existing off-the-shelf TAD models without model retraining.
To maximize the applicability, we consider the TAD inference as a black-box process.
Concretely, taking the predictions by any model,
we formulate the start and end points of action instances
with a Gaussian distribution in a {\em continuous} snippet temporal resolution.
We account for the distribution information of temporal boundaries via
Taylor-expansion based approximation.
This enables TAD inference at {\em sub-snippet} precision (Fig. \ref{fig:subsnip}), creating the possibility of alleviating the temporal quantization error.
We name our method as {\em Gaussian Approximated Post-processing} ({\bf GAP}).

We summarize the {\bf\em contributions} as follows. (I) We identify the previously neglected harming effect of temporal resolution reduction during the pre-processing step in temporal action detection. (II) For the first time, we investigate the resulting temporal quantization error problem from a model generic post-processing perspective. This is realized by modeling the action boundaries with a Gaussian distribution along with an efficient Taylor-expansion based approximation. (III) Extensive experiments show that a wide range of TAD models \cite{lin2019bmn,buch2017sst,lin2018bsn,xu2020g,nag2021few,xu2021boundary,xu2021low} can 
be seamlessly benefited from our proposed GAP method without algorithmic modification and model retraining, achieving the best single model accuracy on THUMOS and ActivityNet.
\textcolor{black}{
Despite this simplicity, the performance improvement obtained from \shortname{} can match those achieved by designing novel models \cite{chen2022dcan}, which is hence significant.}
% is even more significant as unlike many classification tasks, the gains on TAD obtained by SOTA methods published in the past a couple of years are often of similar magnitude as in this paper.}
At the cost of model retraining, our GAP can be integrated with existing TAD models for achieving further gain. Further, our GAP favourably enables lower temporal resolutions for higher inference efficiency with little performance degradation.

\section{Related Works}
\noindent \textbf{Temporal action detection}
\textcolor{black}{
% Substantial progress has been made in TAD. 
Inspired by object detection in static images \cite{ren2016faster},
R-C3D \cite{xu2017r} uses anchor boxes by following the design of {proposal generation and classification}.
With a similar model design, TURN \cite{gao2017turn} aggregates local features to represent snippet-level features for temporal boundary regression and classification. SSN \cite{zhao2017temporal} decomposes an action instance into three stages (starting, course, and ending)
and employs structured temporal pyramid pooling
to generate proposals.
BSN \cite{lin2018bsn} predicts the start, end and actionness at each temporal location and generates proposals with high start and end probabilities. The actionness was further improved in BMN \cite{lin2019bmn} via
additionally generating a boundary-matching confidence map for improved proposal generation. 
GTAN \cite{long2019gaussian}
improves the proposal feature pooling procedure with a learnable Gaussian kernel for weighted averaging. G-TAD \cite{xu2020g}
learns semantic and temporal context via graph convolutional networks for more accurate proposal generation. BSN++ \cite{su2020bsn++} further extends BMN with a complementary boundary generator to capture rich context.
CSA \cite{sridhar2021class} enriches the proposal temporal context via attention transfer. VSGN \cite{zhao2021video} improves short-action localization using a cross-scale multi-level pyramidal architecture. \textcolor{black}{Recently, Actionformer \cite{zhang2022actionformer} and React \cite{shi2022react} proposed a purely DETR based design for temporal action localization at multiple scales.} 
Mostly, existing TAD models suffer from temporal quantization error as the actions are detected in the reduced temporal space. 
We present a model-agnostic post-processing strategy
for generally tackling this problem
% Our GAP is designed to address this limitation by building a refinement module on top of existing models 
without model redesign and retraining at a negligible cost.
}

\noindent \textbf{Temporal boundary refinement} 
methods can designed particularly for improving proposal localization. but still at the snippet level \cite{liu2020progressive,zeng2019graph,lin2021learning,tan2021relaxed,qing2021temporal}.
However, they still perform at the snippet level, and not solve the temporal quantization error problem as we focus on here.
% Although much focus has been directed in improving the detection performance, little notice is given to the temporal resolution reduction. 
% For example,  designed boundary refinement modules to refine coarser action predictions but involves complicated architecture and loss designs. 
Specifically, PGCN \cite{zeng2019graph} modeled the intra-action proposals using graph convolution networks to refine the boundaries. 
% However it is achieved using a complicated network design which is not scalable. 
PBRNet \cite{liu2020progressive} refined the anchor proposals using a two-stage refinement architecture with a complicated loss design. Recent focus has been shifted to anchor-free proposal refinement \cite{lin2021learning,tan2021relaxed,qing2021temporal}
where coarse action proposals are refined using local and global features to obtain fine-grained action proposals. However, the refinement modules are very design specific and cannot be easily adapted to any existing approaches. 
% Moreover, these approaches used transformers \cite{vaswani2017attention} to generate global attention hence not easily deployable. 
AFSD \cite{lin2021learning} used a pyramidal network to generate coarse action proposals and then refined them with boundary pooling based contrastive learning. 
% Large model size and over-complicated loss design make this approach hard to adapt and deploy to embedded systems. 
Very recently, \cite{nag2022gsm} developed a lightweight transformer based proposal-free model with boundary refinement.
Often, large model size and complicated model/loss design are involved in each of these previous methods.
In contrast, we take a completely different perspective (model-agnostic post-processing) and 
solve uniquely the temporal quantization error problem.
% this approach hard to adapt and deploy to embedded systems. %
% Consistent with previous works, this refinement block is also driven by a complicated loss design and added learnable parameters. In contrast to all previous works, we instead investigate
% the issues of post-processing proposal refinement, a largely ignored perspective in the literature. 
% Not only
% do we reveal a big impact of resolution reduction in the process of using temporal representation but also we propose a principled model-agnostic post-processing approach for significantly improving the performance of existing models. 
Crucially, our method can be seamlessly integrated into prior temporal boundary refinement techniques {\em without} complex model redesign.

\textcolor{black}{This work is inspired by 
\cite{zhang2020distribution} tackling human pose estimation in images, a totally different problem compared to more complex TAD we study here. Technically, we make non-trivial contributions by investigating both post-processing and model integration. Also, the temporal boundaries come with start /end pair form, rather than individual human joint keypoints. 
In the literature, human pose estimation and TAD are two independent
research fields with sparse connections.
However, at high level they could share generic challenges such as result post-processing as we focus on here.
%
% with action models rarely related to human pose methods in both design and concept. 
Importantly, post-processing is significantly understudied yet critical as our work reveal for the first time. This is meaningful and insightful to the TAD field,
% going beyond the details of specific techniques. 
a new dimension for temporal action detection as far as we know. 
}

% We consider the largely under-studied {\bf \em temporal quantization error} problem with Temporal Action Detection (TAD). % in a post-processing perspective.
% The {\em objective} is to improve the temporal detection performance of any pre-trained state-of-the-art models {\em without algorithmic modification nor model retraining}.
% To that end, we formulate a novel model-agnostic post-processing approach.
% %
% To facilitate holistic understanding, we start by summarizing existing TAD methods. % into a unified pipeline.

% We consider the temporal snippet representation problem for the refinement of temporal boundaries in untrimmed video. The objective is to predict the start and end points at sub-snippet level which is otherwise affected by the temporal resolution difference between the prediction and the original video. 

% In the following we first describe the temporal action detection problem and the proposal boundary estimation process, focusing on the limitation analysis of the existing standard method and development of a novel solution. 
% In the following we first describe the decoding process, focusing on the limitation analysis of the existing standard method and the development of a novel solution. Then, we further discuss and address the limitations of the encoding process. Lastly, we describe the integration of existing human pose estimation models with the proposed method.
\begin{figure*}[t]
    \centering
    \includegraphics[scale=0.175]{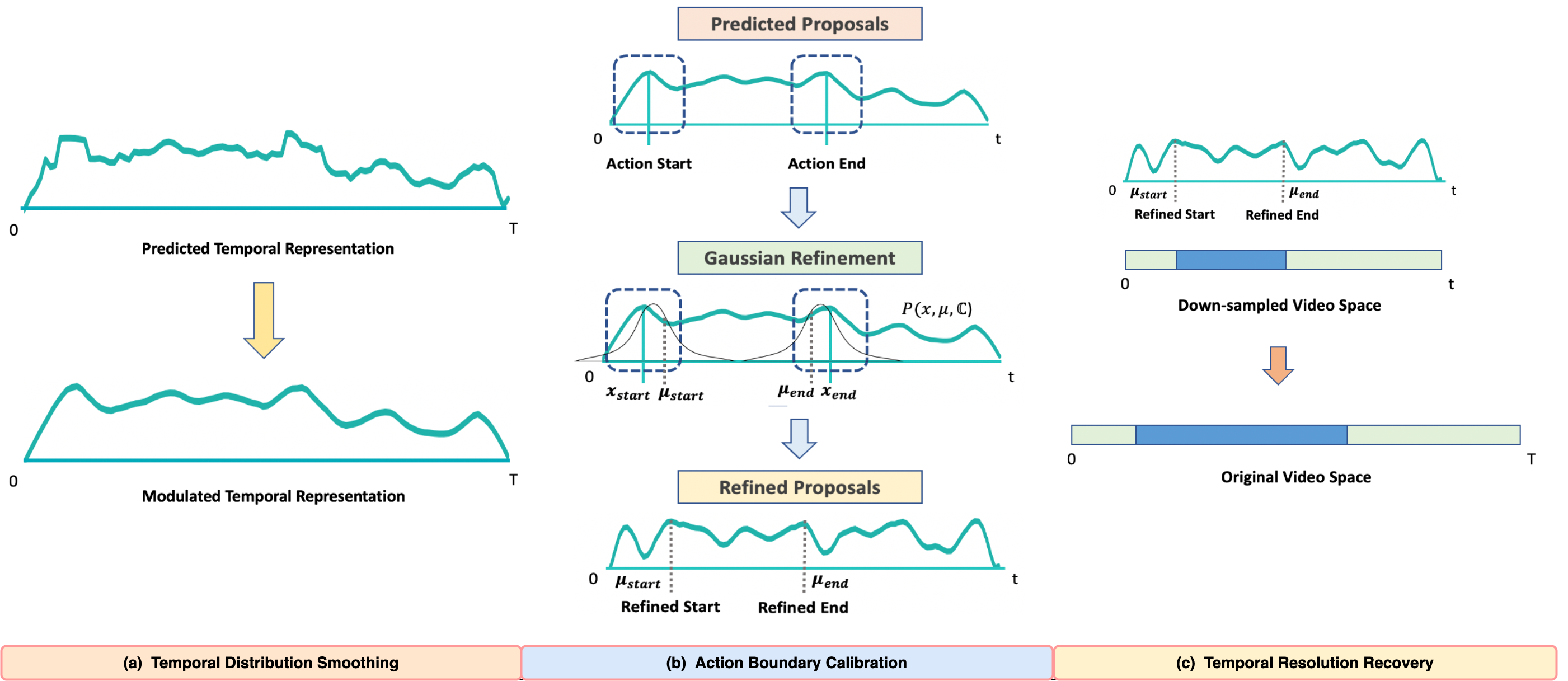}
    % \vspace{0.1in}
    \caption{ \textbf{Overview of the proposed {\em \modelname} (\shortname) method.} Given a test video, an existing TAD model generates 1-D temporal score distribution of candidate foreground action instances. 
    With our \shortname, (a) we first regulate the temporal distribution (Eq~\eqref{eqn_14}) by smoothing the score curve, followed by (b) detecting the boundary (\ie, start/end points) and distributional refinement using a Gaussian kernel to obtain more accurate prediction at the sub-pixel precision.
    % the shifted boundary snippet points for the action proposals. 
    (c) We finally recover the original temporal resolution by multiplying the video-duration with the refined proposals.}
    \label{fig:archi}
    \vspace{-0.15in}
\end{figure*}
\section{Method}
% A TAD model processes any given 

We denote an untrimmed video as $X = \{x_{n}\}_{n=1}^{l_{v}}$ including a total of $l_{v}$ frames.
% , with $x_{n}$ specifying the $n$-th frame. 
Ground-truth annotation of a training video $X_i$ has $M_i$ action instances $\Psi_i = \{(\psi_j, \xi_j, y_j)\}_{j=1}^{M_i}$
where $\psi_{j}$/$\xi_{j}$ denote 
the start/end time, and $y_j$ is the action category.
During both training and inference, any video $V$ is typically {\em pre-processed}
into a unified representation format by first applying a pre-trained, frozen video encoder (\eg, TSN \cite{wang2016temporal}) and then sampling equidistant temporal points for a fixed number of (\eg, 100) times.
Each sampled point is called {\em a snippet} representing a short sequence of consecutive video frames.
Obviously, this pre-processing is a {\bf \em temporal downsampling} procedure,
resulting in TAD 
%
% This makes the following inference by a given TAD model (\eg, GCNext in GTAD \cite{xu2020gtad}) $\phi(.)$ 
at low temporal resolution as:
% and $M_i$ is the action instance number. 
% In this paper, we explicitly focus on temporal action proposal methods as these methods generate action boundary proposals. 
% To generate proposals of input video, first we need to encode visual content of video. In general, any video encoders (\eg, TSN \cite{}) can be used which is pretrained with Kinetics and  are frozen during the training phase. The feature extraction involves sampling equidistant temporal points from the long untrimmed video and then it is passed through a backbone network (\eg 1D Conv) 
% which is followed by localization specific module (\eg GCNext in GTAD \cite{xu2020gtad}) to generate the proposals $\mathbb{P}  = \{s_{i},e_{i}\}_{i=1}^{N_{p}}$ where $s_{i}$ and $e_{i}$ are the predicted start and end times and $N_p$ being the number of action proposals per video. 
\begin{equation}\label{eqn_1}
\mathbb{P} = \phi(F_{v}) = \{s_{i},e_{i}\}_{i=1}^{N_{p}},
\end{equation}
where $s_{i}$ and $e_{i}$ are the start and end time of $i$-th predicted action instance, $N_p$ specifies the number of action predictions per video, and $F_{v}$ denotes the downsampled snippet feature.
To generate the final temporal boundaries, the action predictions $\mathbb{P}$ need to be {\bf \em temporally upsampled} linearly back to the original temporal resolution.

This temporal downsampling and upsampling process introduces
temporal quantization errors negative to model performance.
To address this problem, we propose a model-agnostic {\em Gaussian Approximated Post-processing} (GAP) method as detailed below.
% where $F_{v}$ denotes the extracted video feature and $\phi(.)$ represents the TAD model. 
% Thus predicting the start and end points from a downsampled temporal dimension may include both background and foreground in a single snippet. Hence the snippet prediction scores will reflect such occurance which is utilized during the refinement stage.

\subsection{Temporal Boundary Calibration} 

GAP aims to calibrate the start and end points of a given action boundary prediction. Our key idea is to 
% \noindent\textbf{Our snippet decoding method} 
explore the per-snippet score distribution structure of the predicted proposals $P$ to infer the underlying maximum activation for both the start and end points. 
% This differs dramatically to the
% standard method above relying on a hand-designed offset
% prediction factor $\alpha$, with little design justification and rationale. \\
Specifically, 
% to obtain the accurate location at the degree of sub-snippet, 
we assume the predicted score distribution follows a univariate
Gaussian distribution.
This is conceptually similar with existing TAD methods \cite{lin2019bmn,xu2020g} using the overlap ratio over anchors against the annotated action intervals to create the ground-truth learning objective.
Given a predicted boundary point at a discrete snippet temporal location $x\in [1,2, \cdots, T]$ with $T$ the total number of snippets per video, we formulate the temporal boundary distribution as:
\begin{align}\label{eqn_5} \footnotesize
    P(x;\mu, \mathbb{C}) = \frac{\exp\left( -\frac{1}{2}(x - \mu)^{T}\mathbb{C}^{-1}(x - \mu) \right)}{(2\pi)|\mathbb{C}|^{1/2}},
\end{align}
where $\mathbb{C}$ is the covariance matrix, and $\mu$ refers to the underlying boundary point at sub-snippet resolution.
%In particular, our objective is to reason the value of $\mu$. 
The covariance $\mathbb{C}$ is a diagonal matrix same as used in snippet embedding:
\begin{align}\label{eqn_7}
    \mathbb{C} = \begin{bmatrix}
 \sigma^{2} & 0 \\ 
 0 & \sigma^{2}
\end{bmatrix},
\end{align}
where $\sigma$ is the standard deviation identical for both directions.

In order to reduce the approximation difficulty, we use
logarithm to transform the original exponential form P to a
quadratic form G to facilitate inference while keeping the
original maximum activation location as:

    \begin{align}\label{eqn_8}
    % \begin{gather}
    G(x;\mu,\mathbb{C}) = ln(P) = \\ \nonumber -ln(2\pi)-\frac{1}{2}ln(|\mathbb{C}|) - \frac{1}{2}(x - \mu)^{T}\mathbb{C}^{-1}(x - \mu).
    % \end{gather}
\end{align}

Our objective is to reason the value $\mu$ which refers to the underlying boundary point at sub-snippet resolution. 

As an extreme point in a curve,  it is known that the first derivative at the location $\mu$ meets the condition:
\begin{align}\label{eqn_9}
    \left.\begin{matrix}
 \mathcal{D}^{'}(x)
\end{matrix}\right|_{x=\mu} = 
\left.\begin{matrix} {(\frac{\partial G}{\partial x})^{\top}}\end{matrix} \right|_{x=\mu} = \left. \big(\begin{matrix}
-\mathbb{C}^{-1}(x - \mu)
\end{matrix}\right|_{x=\mu} \big)^{\top} = 0 .
\end{align}

To explore this math condition, we adopt the Taylor’s theorem.
Formally, we approximate the activation $G(\mu)$ by a Taylor
series up to the quadratic term, evaluated at {\em the maximal
activation} $x$ of the predicted snippet distribution as
\begin{align}\label{eqn_10}
    G(\mu) = G(x) + \mathcal{D}^{'}(x)(\mu -x) + \frac{1}{2}(\mu -x)^{T}\mathcal{D}^{''}(x)(\mu - x) 
\end{align}
where $\mathcal{D}^{''}$ is the second derivative (\ie, Hessian)
of $G$ evaluated at $x$, formally defined as:
\begin{align}\label{eqn_11}
    \mathcal{D}^{''}(x) = \frac{\partial \mathcal{D}^{'}(x)}{\partial x} = -\mathbb{C}^{-1}.
\end{align}
% \subsection{Inference}
The intuition is that, $x$ is typically close to the underlying unseen optimal prediction so that the approximation could be more accurate.
% to approximate $\mu$ is that it represents a good coarse joint prediction that approaches $\mu$.
Combining Eq. (\ref{eqn_9}), Eq. (\ref{eqn_10}), Eq. (\ref{eqn_11}) together, we obtain the refined prediction as:
\begin{align}\label{eqn_15}
    \mu = x - ((D^{''}(x))^{-1}D^{'}(x),
\end{align}
where $D^{''}(x)$ and $D^{'}(x)$ can be estimated efficiently from the given score distribution. 
Finally, we use $\mu$ to predict the start and end points in the original video space.
% Once obtaining $\mu$, we replace it in place of $\alpha \Delta s / \alpha \Delta e$  in Eq (\ref{eqn_3}) to predict the shift of the start and ending point and then use  Eq \ref{eqn_4} to predict the start and end points in the original video space. \\

\textit{Discussion}  
% In contrast to the standard method considering the second maximum activation alone in start/end snippet distribution, the
% proposed coordinate decoding fully explores the snippet
% distributional statistics for revealing the underlying maximum more accurately. In theory, our method is based on
% a principled distribution approximation under a training supervision-consistent assumption that the start/end snippets is in a
% Gaussian distribution. 
Our \shortname{} is efficient computationally as it only needs to compute the first and second
derivative of predicted boundary points.
% one pixel location per snippet distribution. 
Existing TAD approaches can be readily
benefited without model redesign and retraining.
% without any computational cost barriers.

\begin{figure}[t]
    \centering
    \includegraphics[scale=0.34]{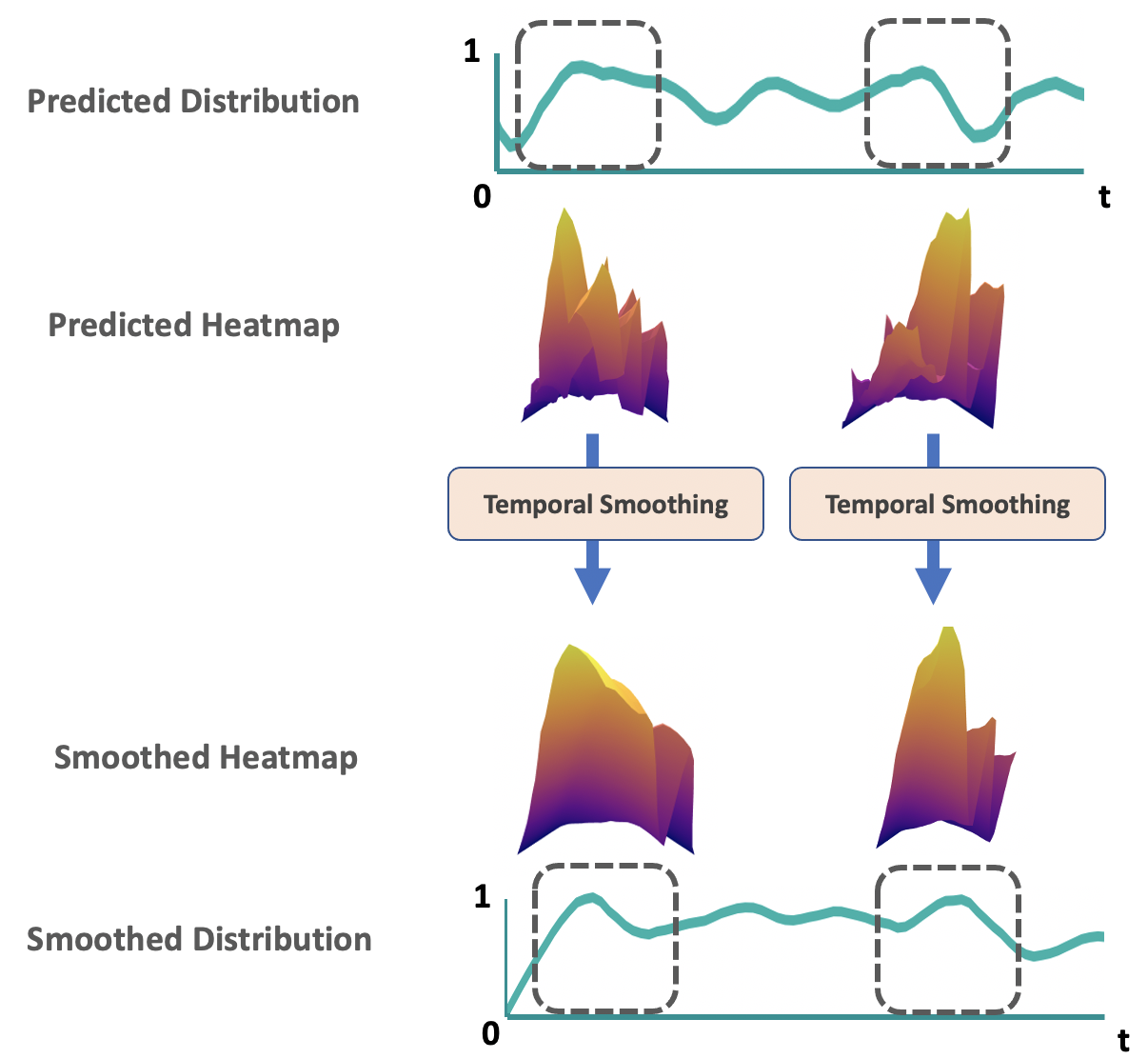}
    \caption{\textbf{Illustration of temporal distribution smoothing} operation along the conflicting action boundary snippets.}
    \vspace{-0.2in}
    \label{fig:smoothing}
\end{figure}

\subsubsection{Temporal distribution smoothing}
Often, the temporal boundary
predicted by a TAD model does not follow
good Gaussian shape. As shown in Fig. \ref{fig:smoothing}, the temporal prediction usually comes with multiple peaks. 
To avoid potential negative effect, we first
smooth the temporal distribution $h$ using a Gaussian kernel $K$ with the same variation as: $h' = K * h$ where $*$ denotes the convolution operation. 
To keep the original magnitude, we further
scale $h'$ linearly as:
% so that its maximum activation is equal to that of $h$ via the following transformation:
\begin{equation}
\label{eqn_14}
    h' = \frac{h' - min(h')}{max(h') - min(h')}*max(h)
\end{equation}
where $max()$ and $min()$ return the maximum and minimum value. 
We validate that this step is useful (Table~\ref{tab:bound}), with the resulting visual effect demonstrated in Fig~\ref{fig:smoothing}.

\subsubsection{Summary} 
Our GAP can be generally integrated with existing boundary regression based TAD models without model redesign and retraining (Fig~\ref{fig:integ}(a)). 
At test time, we take as input the predicted snippet prediction predicted by any model such as BMN, and output more accurate start and end points per prediction in the original video space. 
The pipeline of using \shortname{} is summarized
in Fig. \ref{fig:archi}. 
Totally three steps are involved:
(a) Temporal distribution smoothing (Eq.~\eqref{eqn_14});
(b) Action boundary calibration by Taylor expansion at sub-snippet precision (Eq.~\eqref{eqn_5}-\eqref{eqn_15});
(c) Temporal resolution
recovery linearly to the original video length.

% To address this issue, we simply place the start/end distribution centre at the non-quantised location $g^'$ which represents the accurate ground-truth coordinate. We still apply Eq \ref{eqn_13} but replacing $g^{''}$ with $g^'$. We will demonstrate the benefits of this unbiased snippet distribution generation method (Table xyz).

\subsection{Integration with Existing Model Training}
\begin{figure}[h]
    \centering
    \includegraphics[scale=0.26]{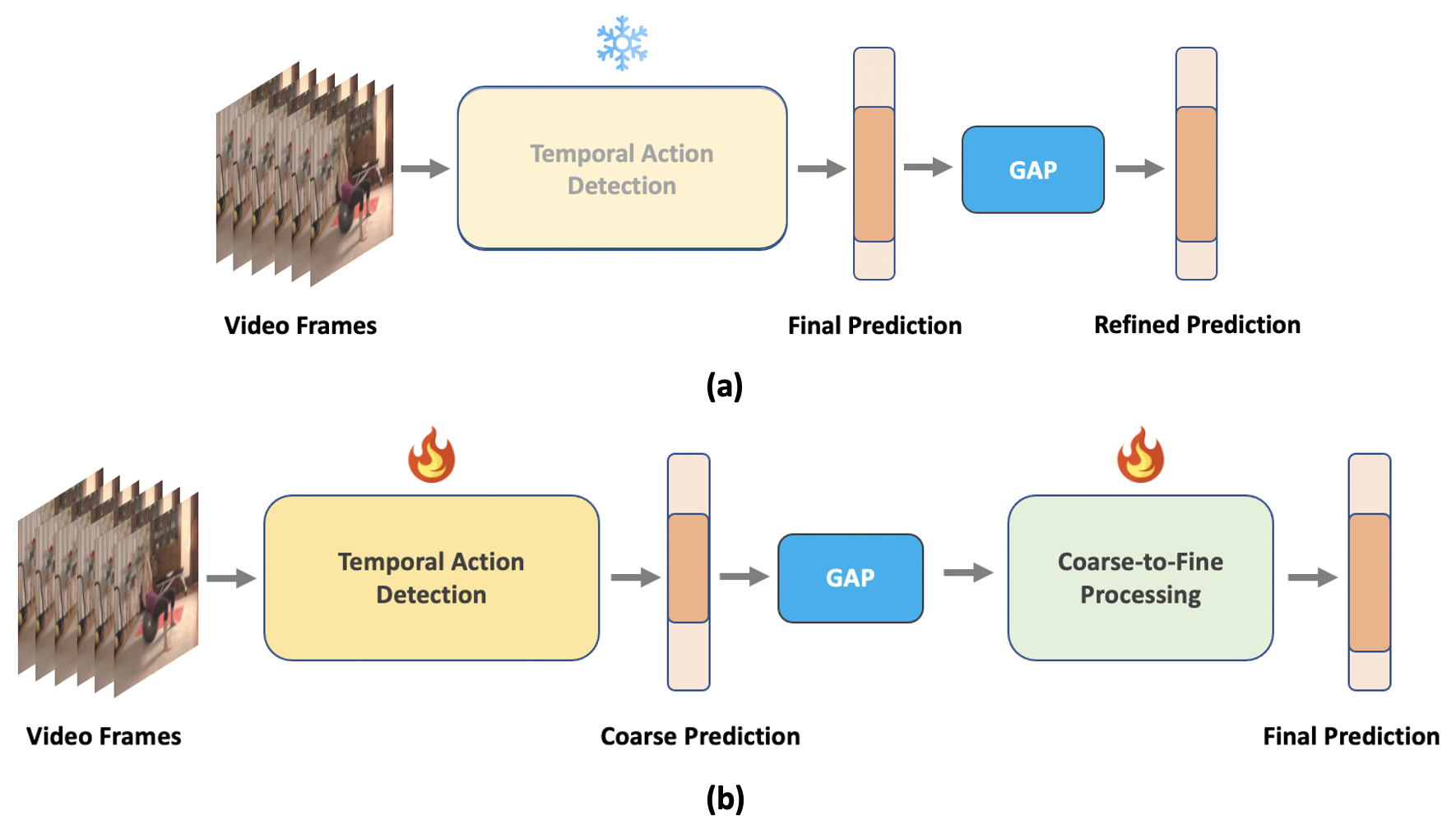}
    \caption{\textbf{Integrating GAP} during (a) post-processing the existing TAD predictions in inference, or (b) model training when applied on intermediate coarse predictions.}
    \vspace{-0.1in}
    \label{fig:integ}
\end{figure}

When model retraining is allowed, our GAP can also be integrated with existing 
TAD training without altering design nor adding learnable parameters (\textcolor{black}{refer to Fig~\ref{fig:integ}(b)}).
The only change is to applying GAP on the intermediate coarser predictions
by prior methods (\eg, AFSD \cite{lin2021learning} and RTDNet \cite{tan2021relaxed}).
\textcolor{black}{While retraining a model with predicted outputs could bring good margin, 
our post processing mode is more generally useful with little extra cost.
% it supports a smaller set of existing methods, while at a cost of retraining. In comparison, our post processing mode is more generally useful with less cost.
}

% a training pipeline when predicting coarser predictions (\eg, in AFSD) and use GAP for feature refinement without adding new parameters. In the whole lifecycle, we keep an existing TAD model intact as the original design. This allows to maximise the generality and scalability of our method.

\begin{table*}[t]
\small
\centering
\setlength{\tabcolsep}{10pt}
\caption{Evaluating the generic benefits of our \shortname{} method on improving state-of-the-art TAD models on the ActivityNetv1.3 and THUMOS14 datasets. \textcolor{black}{Empty results are due to the unavailability of open-source code.}}
\label{tab:main}
\vspace{0.1in}
\begin{tabular}{c|c|cccc|cccc}
\toprule
\multirow{3}{*}{\textbf{Category}} &
  \multirow{3}{*}{\textbf{Method}} &
  \multicolumn{4}{c|}{\textbf{ActivityNet}} &
  \multicolumn{4}{c}{\textbf{THUMOS14}} \\ \cline{3-10} 
                               &        & \multicolumn{4}{c|}{\textbf{mAP}}             & \multicolumn{4}{c}{\textbf{mAP}}               \\ \cline{3-10} 
 &
   &
  \textbf{0.5} &
  \textbf{0.75} &
  \multicolumn{1}{c|}{\textbf{0.95}} &
  \textbf{Avg} &
  \textbf{0.3} &
  \textbf{0.5} &
  \multicolumn{1}{c|}{\textbf{0.7}} &
  \textbf{Avg} \\ \hline
\multirow{4}{*}{Anchor-based}  & MUSES \cite{liu2021multi}  & 50.0 & 34.9 & \multicolumn{1}{c|}{6.5} & 34.0 & 68.9 & 56.9 & \multicolumn{1}{c|}{31.0} & 53.4 \\
                               &  MUSES \cite{liu2021multi} + \textbf{GAP}   &
                               \cellcolor[HTML]{EFEFEF}\textbf{50.3}   & \cellcolor[HTML]{EFEFEF}\textbf{35.5}   & \multicolumn{1}{c|}{\cellcolor[HTML]{EFEFEF}\textbf{6.9}}  & \cellcolor[HTML]{EFEFEF}\textbf{34.3}   & \cellcolor[HTML]{EFEFEF}\textbf{69.3}   & \cellcolor[HTML]{EFEFEF}\textbf{57.8}   & \multicolumn{1}{c|}{\cellcolor[HTML]{EFEFEF}\textbf{31.9}}   & \cellcolor[HTML]{EFEFEF}\textbf{53.8}   \\ \cline{2-10} 
                               & PBRNet \cite{liu2020progressive} & 53.9 & 34.9 & \multicolumn{1}{c|}{8.9} & 35.0 & 58.5 & 51.3 & \multicolumn{1}{c|}{29.5} & -    \\
                               &  PBRNet \cite{liu2020progressive} + \textbf{GAP}   & \cellcolor[HTML]{EFEFEF}\textbf{54.4}   & \cellcolor[HTML]{EFEFEF}\textbf{35.4}   & \multicolumn{1}{c|}{\cellcolor[HTML]{EFEFEF}\textbf{9.2}}  & \cellcolor[HTML]{EFEFEF}\textbf{35.2}   & \cellcolor[HTML]{EFEFEF}\textbf{59.2}   & \textbf{51.9}   & \multicolumn{1}{c|}{\cellcolor[HTML]{EFEFEF}\textbf{30.0}}   & \cellcolor[HTML]{EFEFEF} -   \\ \hline
\multirow{10}{*}{Anchor-Free}  & BMN \cite{lin2019bmn}    & 50.1 & 34.8 & \multicolumn{1}{c|}{8.3} & 33.9 & 56.0 & 38.8 & \multicolumn{1}{c|}{20.5} & 38.5 \\
                               &  BMN \cite{lin2019bmn} + \textbf{GAP}   & \cellcolor[HTML]{EFEFEF}\textbf{50.5}   & \cellcolor[HTML]{EFEFEF}\textbf{35.2}   & \multicolumn{1}{c|}{\cellcolor[HTML]{EFEFEF}\textbf{8.6}}  & \cellcolor[HTML]{EFEFEF}\textbf{34.3}   & \cellcolor[HTML]{EFEFEF}\textbf{56.6}   & \cellcolor[HTML]{EFEFEF}\textbf{39.4}   & \multicolumn{1}{c|}{\cellcolor[HTML]{EFEFEF}\textbf{21.0}}   & \cellcolor[HTML]{EFEFEF}\textbf{38.9}   \\ \cline{2-10} 
                               & GTAD \cite{xu2020boundary}  & 50.4 & 34.6 & \multicolumn{1}{c|}{9.0} & 34.1 & 54.5 & 40.2 & \multicolumn{1}{c|}{23.4} & 39.3 \\
                               &  GTAD \cite{xu2020boundary} + \textbf{GAP}   & \cellcolor[HTML]{EFEFEF}\textbf{50.8}   & \cellcolor[HTML]{EFEFEF}\textbf{34.9}   & \multicolumn{1}{c|}{\cellcolor[HTML]{EFEFEF}\textbf{9.2}}  & \cellcolor[HTML]{EFEFEF}\textbf{34.4}   & \cellcolor[HTML]{EFEFEF}\textbf{55.0}   & \cellcolor[HTML]{EFEFEF}\textbf{40.5}   & \multicolumn{1}{c|}{\cellcolor[HTML]{EFEFEF}\textbf{23.8}}   & \cellcolor[HTML]{EFEFEF}\textbf{39.6}   \\ \cline{2-10} 
                              & DCAN \cite{chen2022dcan} & 51.8 & 35.9 & \multicolumn{1}{c|}{9.4} & 35.4 & 68.2 & 54.1 & \multicolumn{1}{c|}{32.6} & -    \\
                              &  DCAN \cite{chen2022dcan} + \textbf{GAP}   & \cellcolor[HTML]{EFEFEF}\textbf{52.4}   & \cellcolor[HTML]{EFEFEF}\textbf{36.4}   & \multicolumn{1}{c|}{\cellcolor[HTML]{EFEFEF}\textbf{9.6}}  & \cellcolor[HTML]{EFEFEF}\textbf{35.8}   & \cellcolor[HTML]{EFEFEF}\textbf{68.6}   & \cellcolor[HTML]{EFEFEF}\textbf{54.6}   & \multicolumn{1}{c|}{\cellcolor[HTML]{EFEFEF}\textbf{33.0}}   & \cellcolor[HTML]{EFEFEF} -   \\ \cline{2-10} 
                               & RTDNet \cite{tan2021relaxed} & 47.2 & 30.7 & \multicolumn{1}{c|}{8.6} & 30.8 & 68.3 & 51.9 & \multicolumn{1}{c|}{23.7} & -    \\
                               &  RTDNet \cite{tan2021relaxed} + \textbf{GAP}   & \cellcolor[HTML]{EFEFEF}\textbf{47.7}   & \cellcolor[HTML]{EFEFEF}\textbf{31.1}   & \multicolumn{1}{c|}{\cellcolor[HTML]{EFEFEF}\textbf{8.8}}  & \cellcolor[HTML]{EFEFEF}\textbf{31.2}   & \cellcolor[HTML]{EFEFEF}\textbf{68.8}   & \cellcolor[HTML]{EFEFEF}\textbf{52.3 }  & \multicolumn{1}{c|}{\cellcolor[HTML]{EFEFEF}\textbf{24.2}}   & \cellcolor[HTML]{EFEFEF} -   \\ \cline{2-10} 
                               & AFSD \cite{lin2021learning}   & 52.4 & 35.3 & \multicolumn{1}{c|}{6.5} & 34.4 & 67.3 & 55.5 & \multicolumn{1}{c|}{31.1} & 52.0 \\
                               &  AFSD \cite{lin2021learning} + \textbf{GAP}   & \cellcolor[HTML]{EFEFEF}\textbf{53.0}   & \cellcolor[HTML]{EFEFEF}\textbf{35.9}   & \multicolumn{1}{c|}{\cellcolor[HTML]{EFEFEF}\textbf{7.1}}  & \cellcolor[HTML]{EFEFEF}\textbf{34.8}   & \cellcolor[HTML]{EFEFEF}\textbf{68.0}   & \cellcolor[HTML]{EFEFEF}\textbf{56.1}   & \multicolumn{1}{c|}{\cellcolor[HTML]{EFEFEF}\textbf{31.5}}   & \cellcolor[HTML]{EFEFEF}\textbf{52.5}   \\ \cline{2-10}
                               & ActionFormer \cite{lin2021learning}   & 53.5  & 36.2  & \multicolumn{1}{c|}{8.2 } & 35.6 & 82.1  & 71.0  & \multicolumn{1}{c|}{43.9 } & 66.8 \\
                               &  ActionFormer \cite{lin2021learning} + \textbf{GAP}   & \cellcolor[HTML]{EFEFEF}\textbf{53.9}   & \cellcolor[HTML]{EFEFEF}\textbf{36.4}   & \multicolumn{1}{c|}{\cellcolor[HTML]{EFEFEF}\textbf{8.5}}  & \cellcolor[HTML]{EFEFEF}\textbf{36.0}   & \cellcolor[HTML]{EFEFEF}\textbf{82.3}   & \cellcolor[HTML]{EFEFEF}\textbf{71.4}   & \multicolumn{1}{c|}{\cellcolor[HTML]{EFEFEF}\textbf{44.2}}   & \cellcolor[HTML]{EFEFEF}\textbf{66.9}   \\ \cline{2-10}
                               & React \cite{lin2021learning}   & - & - & \multicolumn{1}{c|}{-} & - & 69.2 & 57.1 & \multicolumn{1}{c|}{35.6} & 55.0 \\
                               &  React \cite{lin2021learning} + \textbf{GAP}   & \cellcolor[HTML]{EFEFEF}\textbf{-}   & \cellcolor[HTML]{EFEFEF}\textbf{-}   & \multicolumn{1}{c|}{\cellcolor[HTML]{EFEFEF}\textbf{-}}  & \cellcolor[HTML]{EFEFEF}\textbf{-}   & \cellcolor[HTML]{EFEFEF}\textbf{69.5}   & \cellcolor[HTML]{EFEFEF}\textbf{57.3}   & \multicolumn{1}{c|}{\cellcolor[HTML]{EFEFEF}\textbf{35.7}}   & \cellcolor[HTML]{EFEFEF}\textbf{55.2}   \\
                               
                               \hline
\multirow{2}{*}{Proposal-Free} & TAGS \cite{nag2022gsm}   & 56.3 & 36.8 & \multicolumn{1}{c|}{9.6} & 36.5 & 68.6 & 57.0 & \multicolumn{1}{c|}{31.8} & 52.8 \\
                               &  TAGS \cite{nag2022gsm} + \textbf{GAP}   & \cellcolor[HTML]{EFEFEF}\textbf{56.7}   & \cellcolor[HTML]{EFEFEF}\textbf{37.2}   & \multicolumn{1}{c|}{\cellcolor[HTML]{EFEFEF}\textbf{9.8}}  & \cellcolor[HTML]{EFEFEF}\textbf{36.7}   & \cellcolor[HTML]{EFEFEF}\textbf{69.1}   & \cellcolor[HTML]{EFEFEF}\textbf{57.4}   & \multicolumn{1}{c|}{\cellcolor[HTML]{EFEFEF}\textbf{32.0}}   & \cellcolor[HTML]{EFEFEF}\textbf{53.0}   \\ \hline
\end{tabular}

\end{table*}

\subsubsection{Ground-truth calibration}
\begin{figure}[t]
    \centering
    \includegraphics[scale=0.26]{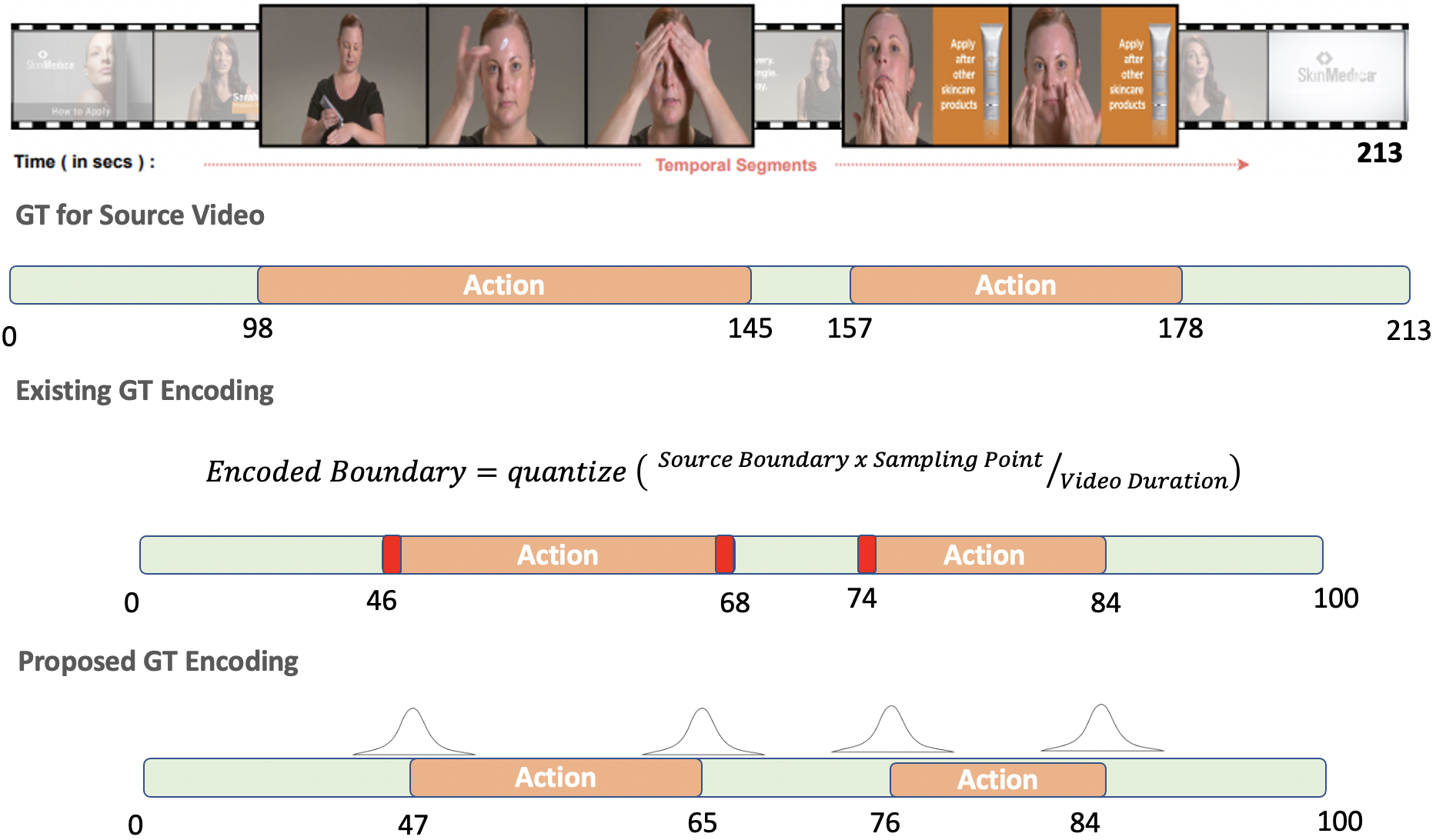}
    \caption{\textbf{Illustration of quantization error} in the standard ground-truth (GT) generation: Obtaining the start/end points with floor based snippet quantization. As a result, an error (indicated by red marker) is introduced. Other quantization (\eg, ceiling, rounding) share the same problem.}
    \label{fig:enc}
    \vspace{-0.15in}
\end{figure}

\textcolor{black}{As model inference,
the ground-truth for training is also affected by temporal resolution reduction.
Specifically, during pre-processing by evenly sampling temporal points from the whole raw video length, the ground-truth start/end snippet locations need to be transformed accordingly.
%
% We have discussed about how temporal resolution reduction affects the overall model performance. The ground-truth calibration also shares the same limitation. 
% Specifically, the standard calibration process starts
% equi-distantly sampling temporal points from the whole raw video length into the model input size. So, the ground-truth start/end snippet locations need to be transformed accordingly.
Formally, we denote the ground-truth of a video as $g = \Psi_g = \{(\hat{s}_j, \hat{e}_j, y_j)\}_{j=1}^{M_i}$ including the start and end annotations of each action instance. The temporal resolution reduction is defined as:
\begin{align}\label{eqn_12}
    g{'} = ({s}{'},{e}{'}) = \frac{g}{\lambda} = (\frac{\hat{s}}{\lambda},\frac{\hat{e}}{\lambda}),
\end{align}
where $\lambda$ is the downsampling parameter conditioned on the temporal sampling ratio and video duration. Conventionally, in the downsampling step, we often quantise $g{'}$: 
\begin{align}\label{eqn_13}
    g{''} = (s{''},e{''}) = quantize(g{'}) = quantize(\frac{\hat{s}}{\lambda},\frac{\hat{e}}{\lambda}),
\end{align}
where $quantize()$ specifies a quantization function (\eg, including floor, ceil and round). \textcolor{black}{Noted that, $g{''}$ is a scalar term which represenets an individual start/end point.}
% individually represents a list of quantized scalar start and end points}. 
Next, the start/end snippet distribution centred at the quantized location $g{''}$ can be synthesized via:
\begin{align}\label{eqn_14}\footnotesize
P(x;g{''}) = \frac{1}{2\pi \sigma^{2}}exp\left(\frac{- (x-g{''})^{2}}{2\sigma^{2}}\right),
\end{align}
where $x$ denotes a point in the temporal distribution and $\sigma$ denotes a fixed spatial variance. This is applied separately on both the ground-truth start and end points. Nonetheless, such start/end snippet distributions generated are clearly inaccurate due to the quantization error, as illustrated in Fig~\ref{fig:enc}. This may cause sub-optimal supervision signals and degraded performance.
% , particularly for accurate temporal boundary predictions as proposed in this work. 
To address this issue, we instead place the start/end centre at the {\em  original non-quantized location} $g{'}$ as it  represents the accurate ground-truth location. Afterwards, we still apply Eq~\eqref{eqn_14} with $g{''}$ replacing by $g'$. We will evaluate the effect of ground-truth calibration (Table~\ref{tab:quant}).
}

% \subsection{Discussion}
% \textcolor{blue}{
% We note that, this work tackles a totally different problem (more complex video action detection vs image human pose detection \citet{zhang2020distribution}). Technically, this is non-trivial as we study both post-processing and model integrating and temporal boundaries come with start /end pair form with action models rarely related to human pose methods in both design and concept. Importantly, post-processing is significantly understudied yet critical as our work shows here for the first time. This is meaningful and insightful to the action detection field, going beyond the details of specific techniques. As this is a new dimension for temporal action detection. 
% }

% The previous section has addressed the problem with snippet decoding, rooted at temporal resolution reduction. 
% Snippet encoding also shares the same limitation.
% Specifically, the standard coordinate encoding method starts
% equidistandtly sampling temporal points from the whole raw video length into the model input size. So, the ground-truth start/end snippet locations need to be transformed accordingly before generating the snippet distribution. 

\section{Experiments}
\label{sec:experiments}

\textbf{Datasets } We conduct extensive experiments on two major TAD benchmarks. (1) ActivityNet-v1.3 \cite{caba2015activitynet} provides 19,994 videos from 200 action classes. We adopt the standard setting to split all the videos into training, validation and testing subsets in a ratio of 2:1:1. (2) THUMOS14 \cite{idrees2017thumos} offers 200 validation videos and 213 testing videos from 20 action categories with labeled temporal boundary and class label. 

\noindent\textbf{Implementation details} We have adopted all the original training and inference details of existing TAD methods. For re-training AFSD \cite{lin2021learning} and RTDNet  \cite{tan2021relaxed}, we have used the reported hyperparameters in the respective papers. All the training has been performed on an Intel i9-7920X CPU with two Nvidia RTX 2080 Ti GPU. We used the same feature encoders as the original papers. During inference, all the full-resolution proposals are passed into SoftNMS for final output similar to \cite{lin2019bmn}. We will release the code upon acceptance.

% Our model is trained for 15 epochs using SGD with learning rate of 10−4/10−5, weight decay of 10−3/10−5 for AcitivityNet/THUMOS respectively. The batch size is set to 100 for ActivityNet and 50 for THUMOS. During testing, we set the threshold set for segmentation mask Θ = {0.1 ∼ 0.9} with step 0.2.

\subsection{Improving State-of-the-Art Methods}
% \usepackage{multirow}

We evaluate the effect of our GAP
on top TAD performers across all the anchor-based, anchor-free and proposal-free methods (MUSES \cite{liu2021multi}, BMN \cite{lin2019bmn}, AFSD \cite{lin2021learning} and TAGS \cite{nag2022gsm})
% Table \ref{tab:main}, shows the accuracy results of the state-of-the-art methods and GAP
on ActivityNet and THUMOS dataset.

\noindent\textbf{Results on ActivityNet }
From Table \ref{tab:main}
we make the following  observations:
{\bf(1)} The performance for anchor-based approaches \cite{liu2021multi,liu2020progressive} is improved by at max 0.3\% in avg mAP and by a constant gain of 0.3\% to 0.5\% in mAP@IOU 0.5. In particular, \shortname{} can further improve over previous offset-based boundary refinement like PBRNet \cite{liu2020progressive}.
{\bf(2)} When applying GAP on anchor-free approaches, the performance gain is in the range from 0.2\% to 0.4\%.
% indicating that fitting gaussian at start and end point helps reduce the quantization error. 
Noticeably, AFSD \cite{lin2021learning} is benefited by an impressive improvement of 0.6\% in mAP@IOU0.5 and 0.4\% in avg mAP. 
This gain is already similar to that ($\sim 0.4\%$) of AFSD's complex
learnable boundary refinement component.
%
% The improvement of 0.5\% is lesser than the learnable boundary refinement gain ($\sim 0.4\%$) proposed in AFSD suggesting the impact our GAP brings. 
% This pattern is similarly observed in RTDNet \cite{tan2021relaxed} which employs an architecture similar to object detection Transformers, indicates GAP is more effective in 1-stage approaches.
%
\textcolor{black}{GAP is also effective for multi-scale DETR based approaches like ActionFormer \cite{zhang2022actionformer} with similar margins achieved on ActivityNet. This gain is consistent with those for anchor-free based models.}
{\bf (3)} With very different masking based architecture design in TAGS \cite{nag2022gsm},
\shortname{} can still consistently yield an improvement of 0.2\% in avg mAP.
This further validates the model-agnostic advantage of our method.

% Proposal-free approaches shows a improvement of 0.2\% in avg mAP when GAP is used as a post-processing unit. However, it must be noted that although TAGS \cite{nag2022gsm} is 1-stage, the proposal-free approaches do not have separate start/end point regressors. Hence, we fit the gaussian distribution at the mask start/end point. This validates that GAP is model-agnostic and can bring in improvement in any of anchor-based, anchor-free and proposal-free designs without any change in model design. 

% % \vspace{0.1cm}
\noindent \textbf{Results on THUMOS14 } 
Overall, similar conclusions can be drawn on THUMOS. 
All the models with our proposed GAP post-processing achieve the best results, often by a margin of 0.2$\sim$0.5\% in avg mAP. There is a  noticeable difference that the improvement by GAP is more significant than on ActivityNet, 
% on all the TAD models 
indicating the more severe quantization error on THUMOS due to longer videos.

\noindent
\textcolor{black}{{\bf Discussion } We note that while TAD performance is saturating and a very challenging metric (average mAP over IoU thresholds from 0.5 to 0.95 for ActivityNet and from 0.3 to 0.7 for THUMOS) is applied, \shortname{} can still push the performance at the comparable magnitude as recent state-of-the-art model innovation \cite{chen2022dcan}. This is encouraging and meaningful, except neglectable cost added and no model retraining.
% the improvement from \shortname{} is significant in the community and shows similar/better improvement than major recent state-of-the-art methods.
}
% . This is obvious since the dataset characteristics are different; THUMOS has larger video duration thus $V_{d}$ is more and hence smaller shift in reduced dimesnion has larger shift in original video space. 
% (2) The performance gain of TAD models using GAP is significant in IOU@0.5 and IOU@0.7 consistently, verifying the effectiveness of solving the quantization error.

% \textcolor{black}{When using TS features, {\shortmodelname} achieves again the best results, beating strong competitors like TCANet \cite{wang2021temporal}, CSA \cite{sridhar2021class} by a clear margin. % of xx\%.
% There are some noticeable differences:
% (1) We find that I3D is now much more effective than two-stream (TS),
% \eg, 8.8\% gain in average mAP over TS with {\shortmodelname}, compared with 0.6\% on ActivityNet. This is mostly likely caused by the distinctive characteristics of the two datasets in terms of action instance duration and video length.
% (2) Our method achieves the second best result with marginal edge behind MUSES \cite{liu2021multi}.
% % in only 0.6 and 0.7 mAP. 
% This is partly due to that MUSES benefits from additionally tackling the scene-changes. \textcolor{black}{(3) Our model achieves the best results in stricter IOU metrics (\eg, IOU@0.5/0.6/0.7) consistently using both TS and I3D features, verifying the effectiveness of solving mask redundancy.}
% }

\begin{table}[]
    \centering
    \caption{Effect of temporal size on the ActivityNet using BMN \cite{lin2019bmn} model.}
\vspace{0.1in}
% \resizebox{\textwidth}{!}{
\begin{tabular}{c|c|cccc}
\hline
\multirow{2}{*}{\textbf{Method}} &
  \multirow{2}{*}{\textbf{\begin{tabular}[c]{@{}c@{}} Temporal \\ Resolution\end{tabular}}} &
  \multicolumn{4}{c}{\textbf{mAP}} \\ \cline{3-6} 
     &     & \textbf{0.5} & \textbf{0.75} & \multicolumn{1}{c|}{\textbf{0.95}} & \textbf{Avg} \\ \hline
BMN  & 25  & 44.7           & 27.9            & \multicolumn{1}{c|}{7.0}            & 28.1           \\
BMN+\textbf{GAP} & 25  & \textbf{45.5}           & \textbf{28.4}            & \multicolumn{1}{c|}{\textbf{7.3}}            & \textbf{28.5}           \\ \hline
BMN  & 100 & 50.1         & 34.8          & \multicolumn{1}{c|}{8.3}           & 33.9         \\
BMN+\textbf{GAP} & 100 & \textbf{50.5}           & \textbf{35.2}            & \multicolumn{1}{c|}{\textbf{8.6}}            & \textbf{34.3}           \\ \hline
BMN  & 400 & 50.9           & 34.9            & \multicolumn{1}{c|}{8.1}            & 34.0           \\
BMN+\textbf{GAP} & 400 & \textbf{51.1}           & \textbf{35.0}            & \multicolumn{1}{c|}{\textbf{8.2}}            & \textbf{34.1}           \\ \hline
\end{tabular}
% }
\label{tab:size}
\end{table}

\begin{table}[]
    \centering
    \caption{Effect of temporal smoothing on ActivityNet}
\vspace{0.1in}
% \resizebox{\textwidth}{!}{
\begin{tabular}{c|c|cc}
\hline
\multirow{2}{*}{\textbf{Method}} & \multirow{2}{*}{\textbf{Smoothing}} & \multicolumn{2}{c}{\textbf{mAP}} \\ \cline{3-4} 
                        &                     & \textbf{0.5}        & \textbf{Avg}        \\ \hline
BMN \cite{lin2019bmn}                     & -                   & 50.1       & 33.9       \\ \hline
BMN +\textbf{GAP}                 & \xmark                  & 50.3         & 34.0         \\
BMN +\textbf{GAP}                 & \cmark                 & \textbf{50.5}       & \textbf{34.3}       \\ \hline
\end{tabular}
% }
\label{tab:bound}
\end{table}

\begin{figure*}[t]
    \centering
    \includegraphics[scale=0.40]{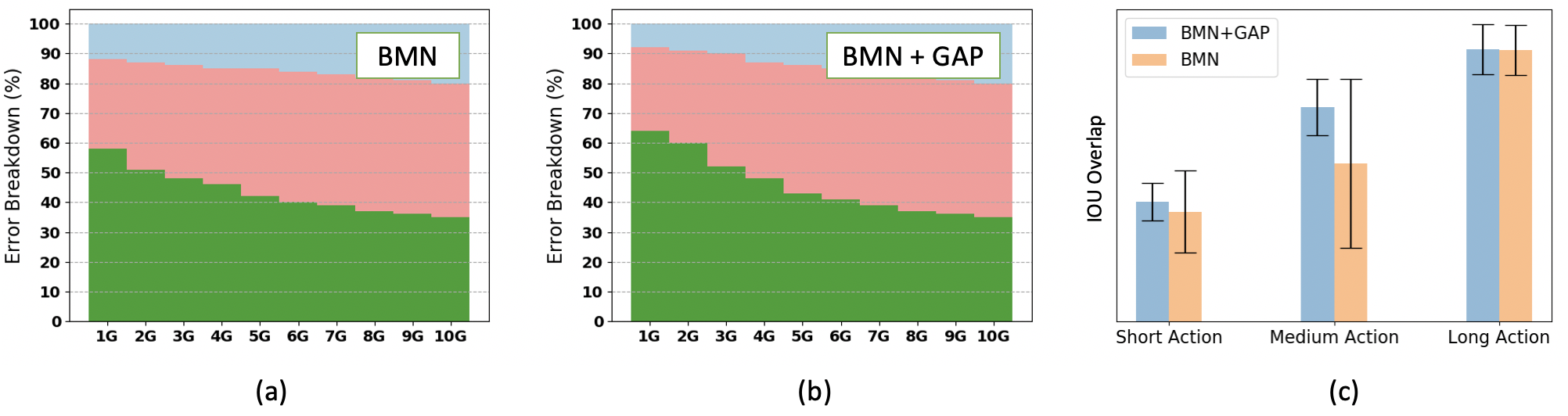}
    \vspace{0.09in}
    \caption{(a) False positive profile of BMN. (b) BMN with GAP on ActivityNet. (c) Proposal overlap analysis on various video lengths.
  We use top up-to 10$G$ predictions per video, where $G$ is the number of ground-truth action instances.}
    \label{fig:error_prof}
    % \vspace{-0.4in}
\end{figure*}
\subsection{Ablation Studies}
\noindent\textbf{(i) Input temporal resolution}
We examined the impact of snippet temporal resolution/size,
% by testing a number of different number of temporal dimesnion, 
considering that it is an important efficiency factor.
% relevant to model inference efficiency. 
We used BMN \cite{lin2019bmn} as the baseline TAD model
% We compared our GAP model
% (BMN \cite{lin2019bmn} as backbone) with the original BMN \cite{lin2019bmn}
% using 
in the standard training and testing setting. From Table~\ref{tab:size} we have a couple of observations: (a) When reducing the input temporal resolution,
as expected the model performance consistently degrades
whilst the inference cost drops. (b) With the support
of GAP, the model performance loss can be effectively
mitigated, especially at very small input resolution. This facilitates the deployment of TAD models on low-resource devices as desired in emerging embedded AI.

\noindent\textbf{(ii) Effect of temporal smoothing} 
We evaluated the effect of temporal smoothing.
% on the overall performance of GAP. 
From the results in Table~\ref{tab:bound}, it can be observed that this step is useful and necessary otherwise the original prediction scores
are less compatible with our \shortname{}.

\begin{table}[t]
        % \begin{minipage}{.42\linewidth}
    % \setlength{\tabcolsep}{5pt}
    \centering
\caption{Speed analysis of existing TAD method w/ our GAP on a NVIDIA RTX 2080 Ti GPU}
\vspace{0.1in}
% \resizebox{\textwidth}{!}{
\begin{tabular}{c|c|c}
\hline
\textbf{Method} & \textbf{\begin{tabular}[c]{@{}c@{}}Inference \\ Time\end{tabular}} & \textbf{\begin{tabular}[c]{@{}c@{}} Speed\end{tabular}} \\ \hline
AFSD \cite{lin2021learning}            & 0.29 sec                                                           & 1931 FPS                                                            \\
AFSD + \textbf{GAP}      & 0.31 sec                                                           & 1792 FPS                                                            \\ \hline
\end{tabular}
% }
\label{tab:speed}
\end{table}

\noindent\textbf{(iii) Error sensitivity analysis}
We compare our GAP (with BMN backbone) with original BMN \cite{lin2019bmn} \textcolor{black}{(anchor-free)} via false positive analysis \cite{alwassel2018diagnosing}. We sort the predictions by the scores and take the top-scoring predictions per video. Two major errors of TAD are considered:
\textcolor{black}{(1) {\em Localization error}, which is defined as when a proposal/mask is predicted as foreground, has a minimum tIoU of 0.1 but does not meet the tIoU threshold. 
(2) {\em Background error}, which happens when a proposal/mask
is predicted as foreground but its tIoU with ground truth instance is smaller than 0.1.
In this test, we use ActivityNet. 
\textcolor{black}{We observe in Fig. \ref{fig:error_prof}(a,b) that GAP has the most true positive samples at every
amount of predictions. The proportion of localization error
with GAP is also notably smaller,
which is the most critical metric for improving average mAP \cite{alwassel2018diagnosing}. Also based on various video lengths \cite{alwassel2018diagnosing}, we estimated the standard deviation of all the proposal overlap with GT for both BMN and BMN with GAP variant. From Fig~\ref{fig:error_prof}(c), it is interesting to note that our GAP indeed improves the overlap in challenging short and medium length videos and also BMN has a significant standard deviation in shorter-action instances. This explains the gain of GAP refinement over existing BMN which is caused due to the quantization error.}}

\noindent\textbf{(iv) Complexity}
We tested the inference efficiency impact
by our method in AFSD at input size of 100 snippets for ActivityNet on a machine with one i9-7920X CPU and one RTX 2080 GTX
GPU. From Table ~\ref{tab:speed} it can be observed that the running speed is reduced from 1931 FPS to 1792 FPS in the low-efficient python environment, \ie, a drop of 7.2\%. There is a minor affordable increase from post-processing.
% ($\sim$ 6.8\%).
% Hence, the extra cost from GAP is rather affordable. 
Other programming language (\eg, C/C++) based
software can further reduce the overhead addition.

\noindent\textbf{(v) Ground-truth calibration}
\textcolor{black}{We tested the effect of our ground-truth calibration. 
We considered both cases with and without GAP post-processing.
% our proposed ground-truth calibration with the standard calibration, along with both the standard and our boundary calibration method in the inference-only GAP variant. 
We observed from Table~\ref{tab:enc} that our ground-truth calibration brings positive performance margin consistently.
% regardless of the temporal boundary calibration.  
In particular, it contributes consistently a gain of around 0.1\% in avg mAP in both cases particularly for the stricter IOU metrics.
This is reasonable since such fine-grained tuning matters most 
to more demanding metrics.
% suggesting the need of such calibration.
% which is again neglected by previous investigations.
}

% the inference speed.

\noindent\textbf{(vi) Quantization function} \textcolor{black}{We evaluated the quantization function in ground-truth calibration (Eq~\eqref{eqn_13}). Common quantization options include \texttt{floor}, \texttt{ceiling} and \texttt{rounding}. 
% We ablate the impact for each of such techniques in this experiment. 
From Table~\ref{tab:quant} we observed that \texttt{rounding} and \texttt{floor} are similarly effective, whilst \texttt{ceiling} gives the worst performance with a drop of 0.3\% in avg mAP.}

\begin{table}[t]
\centering
\caption{Effect of ground-truth calibration on ActivityNet.
% validation set. 
{\em TAD model}: BMN \cite{lin2019bmn} w/ GAP.
% The model used here is BMN \cite{lin2019bmn} 
}
\begin{tabular}{c|c|cc}
\hline
\multirow{2}{*}{\textbf{Ground-truth}} & \multirow{2}{*}{\textbf{Post-processing}} & \multicolumn{2}{c}{\textbf{mAP}} \\ \cline{3-4} 
                                 &                                  & \textbf{0.5}    & \textbf{Avg}   \\ \hline
W/O Calibration                           & W/O GAP                         & 50.1              & 33.9             \\
W/ Calibration                         & W/O GAP                         & 50.2              & 34.0             \\ \hline
W/O Calibration                           & W/ GAP                             &\bf 50.5              & 34.2             \\
W/ Calibration                         & W/ GAP                             & \textbf{50.5}              & \textbf{34.3}             \\ \hline
\end{tabular}
\label{tab:enc}
\end{table}

\begin{table}[]
\centering
\caption{Effect of ground-truth quantization on ActivityNet.
{\em TAD model}: BMN \cite{lin2019bmn} w/ GAP.}
\begin{tabular}{c|cccc}
\hline
\multirow{2}{*}{\textbf{Quantization Type}} & \multicolumn{4}{c}{\textbf{mAP}}                             \\ \cline{2-5} 
                                            & \textbf{0.5}  & 0.75          & 0.95         & \textbf{Avg}  \\ \hline
Ceiling                                     & 50.2          & 34.9          & 8.2          & 34.0          \\
Rounding                                    & \textbf{50.6} & 35.1          & 8.4          & 34.2          \\
Floor                                       & 50.5          & \textbf{35.2} & \textbf{8.6} & \textbf{34.3} \\ \hline
\end{tabular}
\label{tab:quant}
\vspace{-0.15in}
\end{table}

% \begin{table}[]
% \caption{Speed analysis}
% \vspace{0.1in}
% \begin{tabular}{c|c|c}
% \hline
% \textbf{Method} & \textbf{\begin{tabular}[c]{@{}c@{}}Inference \\ Time\end{tabular}} & \textbf{\begin{tabular}[c]{@{}c@{}} Speed\end{tabular}} \\ \hline
% AFSD            & 0.29 sec                                                           & 1931 FPS                                                            \\
% AFSD + GAP      & 0.31 sec                                                           & 1792 FPS                                                            \\ \hline
% \end{tabular}
% \label{tab:speed}
% \end{table}
% \begin{table}[]
% \begin{tabular}{c|c|c|c}
% \hline
% Level         & Method        & \# Param & \# FLOPS \\ \hline
% No Refinement & TCANet        & 11       & 12       \\ \hline
% Inference     & TCANet + Ours & 21       & 22       \\
% Feature       & TCANet + Ours & 31       & 32       \\ \hline
% \end{tabular}
% \end{table}

% \begin{figure}[]
%     \centering
%     \includegraphics[height=1.2in,width=\columnwidth]{example-image-b}
%     \caption{Caption}
%     \label{fig:my_label}
% \end{figure}
\begin{table}[]
\centering
\caption{Results of integrating GAP in training and inference on ActivityNet}
\begin{tabular}{c|cc|c|cc}
\hline

\multirow{2}{*}{\textbf{Method}}                               & \multicolumn{2}{c|}{\textbf{GAP Integration}}                        & \multirow{2}{*}{\textbf{FLOPS}} & \multicolumn{2}{c}{\textbf{mAP}} \\ \cline{2-3} \cline{5-6} 
                                                               & \multicolumn{1}{c|}{\textbf{Train}} & \textbf{Test} &                                 & \textbf{0.5}    & \textbf{Avg}   \\ \hline
\multirow{3}{*}{RTDNet} & \multicolumn{1}{c|}{\xmark}                   & \xmark                      & \multirow{3}{*}{85.7}           & 47.2            & 30.8           \\ \cline{2-3} \cline{5-6} 
                                                               & \multicolumn{1}{c|}{\cmark}                  & \xmark                       &                                 & 47.8            & 31.4           \\
                                                               & \multicolumn{1}{c|}{\cmark}                  & \cmark                      &                                 & \textbf{47.9}              & \textbf{31.5}             \\ \hline
\multirow{3}{*}{AFSD}  & \multicolumn{1}{c|}{\xmark}                   & \xmark                       & \multirow{3}{*}{157.1}          & 52.4            & 34.4           \\ \cline{2-3} \cline{5-6} 
                                                               & \multicolumn{1}{c|}{\cmark}                  & \xmark                       &                                 & 53.2            & 35.0           \\
                                                               & \multicolumn{1}{c|}{\cmark}                  & \cmark                      &                                 & \textbf{53.4}              & \textbf{35.1}             \\ \hline
\end{tabular}
\label{tab:refine}
\end{table}

\subsection{Integrating GAP with model training}
\begin{figure}[h]
    \centering
    \includegraphics[scale=0.42]{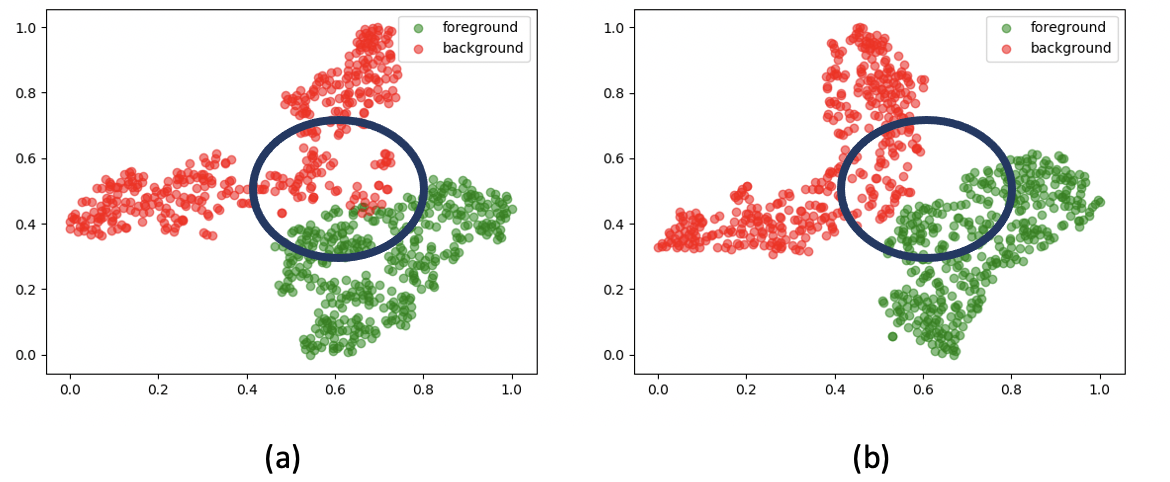}
    \caption{\textbf{T-SNE visualization} of the feature representation of a random ActivityNet val video (a) without and (b) with GAP assisted training. As seen from the encircled region that the original TAD model suffers from the ambiguous boundaries between action foreground and background.  
    % re-trainctioning the conflicting foregrounds are falsely classified as background and vice-versa. 
    This can be well resolved once GAP is integrated during training.}
    \label{fig:feat}
\end{figure}

Other than post-processing,
our GAP can also be integrated into the training of existing TAD models.
We experimented AFSD \cite{lin2021learning} and RTDNet \cite{tan2021relaxed} by applying GAP to their 
intermediate coarse start/end points during training.
Table ~\ref{tab:refine} shows that GAP can bring in more significant gains of 0.7\%$\sim$1.0\% in IOU@0.5 mAP without adding extra parameters nor loss design complexity. \textcolor{black}{This is also clearly reflected in the feature visualization as shown in Fig~\ref{fig:feat}, where the previously ambiguous boundaries between action foreground and background can be well separated.} This suggests more promising benefit of our \shortname{} when model re-training is allowed. \textcolor{black}{We also observed additional gain when integrating \shortname{} during both training and post-processing, indicating flexible usage of our proposed \shortname{} in existing TAD models.}

\section{GAP in different settings}
Due to the nature of being a plug-and-play module, our proposed GAP can be readily used with any Temporal Action Detection (TAD) frameworks irrespective of the supervision setting.
\subsection{GAP in Semi-Supervised Setting}
We integrate the proposed GAP to state-of-the-art semi-supervised TAD approaches. In this experiment, we test on 10\% unlabeled data setting on ActivityNet dataset using two representative semi-supervised approaches: SSTAP \cite{wang2021self}, and SPOT \cite{nag2022pftm}. Since SPOT \cite{nag2022pftm} has 2-stage training (pre-training then finetune), we use GAP once in pre-training and once during inference. It is to be noted that since 
When GAP is used for unsupervised pre-training, %we use ground-truth modulation on the pseudo-ground truth. 
we apply the modulation on the pseudo-ground truth.
From the results in Table~\ref{tabsup:2} it is evident that GAP indeed brings improvement of 0.2$\sim$0.4 \% in avg mAP when used during inference. This indicates that in case of few-labeled data, the detection is inferior which can be improved to some extent using GAP. When used during training in SPOT \cite{nag2022pftm}, it shows further improved performance of 0.7\% indicating that quantization error can be curbed during pre-training time.

\begin{table}[h]
\centering
\small
\setlength{\tabcolsep}{10pt}
\caption{Effect of our GAP on semi-supervised methods on ActivityNet dataset using 10\% labeled data setting.}
\label{tabsup:2}
\begin{tabular}{l|cccc}
\hline
                                  & \multicolumn{4}{c}{\textbf{mAP}}                                                                                                       \\ \cline{2-5} 
\multirow{-2}{*}{\textbf{Method}} & \textbf{0.5}               & \textbf{0.75}              & \multicolumn{1}{c|}{\textbf{0.95}}              & \textbf{Avg}               \\ \hline
SSTAP \cite{wang2021self}                              & 40.7                           & 29.6                          & \multicolumn{1}{c|}{9.0 }                         & 28.2                         \\ 
SSTAP \cite{wang2021self} + \textbf{GAP}                              & \cellcolor[HTML]{FFFFC7}\textbf{41.5}                         & \cellcolor[HTML]{FFFFC7}\textbf{30.2}                         & \multicolumn{1}{c|}{\cellcolor[HTML]{FFFFC7}\textbf{9.1}}                         & \cellcolor[HTML]{FFFFC7}\textbf{28.6}                         \\ \hline
SPOT \cite{nag2022pftm}                              & 49.9                          & 31.1                          & \multicolumn{1}{c|}{8.3 }                         & 32.1                         \\ \hline
                                  & \multicolumn{4}{c}{Training}                                                                                                 \\ \cline{2-5} 
                                  & \cellcolor[HTML]{DAE8FC}\textbf{52.8} & \cellcolor[HTML]{DAE8FC}\textbf{31.6} & \multicolumn{1}{c|}{\cellcolor[HTML]{DAE8FC}\textbf{8.8}} & \cellcolor[HTML]{DAE8FC}\textbf{32.8} \\ \cline{2-5} 
                                  & \multicolumn{4}{c}{Inference}                                                                                                    \\ \cline{2-5} 
\multirow{-4}{*}{SPOT \cite{nag2022pftm}  + \textbf{GAP}}      & \cellcolor[HTML]{FFFFC7}\textbf{52.3} & \cellcolor[HTML]{FFFFC7}\textbf{31.4} & \multicolumn{1}{c|}{\cellcolor[HTML]{FFFFC7}\textbf{8.5}} & \cellcolor[HTML]{FFFFC7}\textbf{32.3} \\ \hline
\end{tabular}
\end{table}

\subsection{GAP in Weakly-Supervised Setting}
We evaluate the effect of our GAP with top performing weakly-supervised TAD methods including popular approaches like DELU \cite{chen2022dual}, CoLA \cite{zhang2021cola} and ASL \cite{ma2021weakly}. 
This test is done on THUMOS14 dataset.
Similar to supervised TAD approaches, as shown in Table \ref{tabsup:3} weakly-supervised methods greatly benefit from GAP during post-processing.

\begin{table}[h]
\centering
\small
\caption{Effect of our GAP on weakly-supervised methods on THUMOS dataset.}
\label{tabsup:3}
\setlength{\tabcolsep}{5pt}
\begin{tabular}{l|cccccc}
\hline
                                 & \multicolumn{6}{c}{\textbf{mAP}}                                                                                                                                                                                                                       \\ \cline{2-7} 
\multirow{-2}{*}{\textbf{Model}} & \textbf{0.3}                        & \textbf{0.4}                        & \textbf{0.5}                        & \textbf{0.6}                        & \multicolumn{1}{c|}{\textbf{0.7}}                        & \textbf{Avg}                        \\ \hline
ASL  \cite{ma2021weakly}                            & 51.8                                   & -                                  & 31.1                                   & -                                   & \multicolumn{1}{c|}{11.4}                                  & 32.2                                  \\
ASL \cite{ma2021weakly} + \textbf{GAP}               & \cellcolor[HTML]{FADDDB}\textbf{53.0} & \cellcolor[HTML]{FADDDB}\textbf{-} & \cellcolor[HTML]{FADDDB}\textbf{31.7} & \cellcolor[HTML]{FADDDB}\textbf{-} & \multicolumn{1}{c|}{\cellcolor[HTML]{FADDDB}\textbf{11.5}} & \cellcolor[HTML]{FADDDB}\textbf{32.4} \\ \hline
CoLA \cite{zhang2021cola}                             &  51.5                                   & 41.9                                   & 32.2                                   & 22.0                                   & \multicolumn{1}{c|}{13.1 }                                  & 40.9                                  \\
CoLA \cite{zhang2021cola} + \textbf{GAP}              & \cellcolor[HTML]{FADDDB}\textbf{51.8} & \cellcolor[HTML]{FADDDB}\textbf{42.2} & \cellcolor[HTML]{FADDDB}\textbf{32.4} & \cellcolor[HTML]{FADDDB}\textbf{22.2} & \multicolumn{1}{c|}{\cellcolor[HTML]{FADDDB}\textbf{13.2}} & \cellcolor[HTML]{FADDDB}\textbf{41.0} \\ \hline
TS-PCA \cite{liu2021blessings}                           & 52.4                                   & 43.5                                   & 34.6                                   & 23.7                                   & \multicolumn{1}{c|}{12.6}                                  & -                                  \\
TS-PCA \cite{liu2021blessings} + \textbf{GAP}            & \cellcolor[HTML]{FADDDB}\textbf{52.9} & \cellcolor[HTML]{FADDDB}\textbf{44.0} & \cellcolor[HTML]{FADDDB}\textbf{34.9} & \cellcolor[HTML]{FADDDB}\textbf{24.0} & \multicolumn{1}{c|}{\cellcolor[HTML]{FADDDB}\textbf{12.8}} & \cellcolor[HTML]{FADDDB}\textbf{-} \\ \hline
CO2-Net \cite{hong2021cross}                          &  54.5                                   & 45.7                                   & 38.3                                   & 26.4                                  & \multicolumn{1}{c|}{13.4 }                                  & -                                  \\
CO2-Net \cite{hong2021cross} + \textbf{GAP}           & \cellcolor[HTML]{FADDDB}\textbf{54.9} & \cellcolor[HTML]{FADDDB}\textbf{46.0} & \cellcolor[HTML]{FADDDB}\textbf{38.8} & \cellcolor[HTML]{FADDDB}\textbf{27.1} & \multicolumn{1}{c|}{\cellcolor[HTML]{FADDDB}\textbf{14.0}} & \cellcolor[HTML]{FADDDB}\textbf{-} \\ \hline
ASM-Loc \cite{he2022asm}                          & 57.1                                   & 46.8                                   & 36.6                                   & 25.2                                   & \multicolumn{1}{c|}{13.4}                                  & 45.1                                  \\
ASM-Loc \cite{he2022asm}  + \textbf{GAP}           & \cellcolor[HTML]{FADDDB}\textbf{58.1} & \cellcolor[HTML]{FADDDB}\textbf{47.5} & \cellcolor[HTML]{FADDDB}\textbf{37.1} & \cellcolor[HTML]{FADDDB}\textbf{25.6} & \multicolumn{1}{c|}{\cellcolor[HTML]{FADDDB}\textbf{13.8}} & \cellcolor[HTML]{FADDDB}\textbf{45.5} \\ \hline
DELU \cite{chen2022dual}                             & 56.5                                   & 47.7                                   & 40.5                                   & 27.2                                   & \multicolumn{1}{c|}{15.3}                                  & 46.4                                  \\
DELU \cite{chen2022dual} + \textbf{GAP}              & \cellcolor[HTML]{FADDDB}\textbf{57.0} & \cellcolor[HTML]{FADDDB}\textbf{48.1} & \cellcolor[HTML]{FADDDB}\textbf{40.9} & \cellcolor[HTML]{FADDDB}\textbf{27.6} & \multicolumn{1}{c|}{\cellcolor[HTML]{FADDDB}\textbf{15.5}} & \cellcolor[HTML]{FADDDB}\textbf{46.6} \\ \hline
\end{tabular}
\end{table}

\subsection{GAP in Few-Shot Setting}
Our GAP can also be used in few-shot temporal action detection approaches. For this experiment, we evaluate our GAP using a recent few-shot TAD approach QAT \cite{nag2021few}. We report 1/5-shot experiment result on ActivityNet. From Table~\ref{tabsup:4}, it is evident that GAP brings largest avg mAP improvement of 0.6\% in 1-shot setting, indicating that the quantization error is high when there are very few labeled samples. 
This error reduces as we increase the number of shots, as expected.
%
% which is trivial. 

\begin{table}[h]
\centering
\small
\caption{Effect of our GAP on few-shot action detection methods on ActivityNet dataset in 1-way multi-instance setting.}
\label{tabsup:4}
\begin{tabular}{c|l|cccc}
\hline
                                &                                   & \multicolumn{4}{c}{\textbf{mAP}}                                                                                                                                           \\ \cline{3-6} 
\multirow{-2}{*}{\textbf{Shot}} & \multirow{-2}{*}{\textbf{Models}} & \textbf{0.5}                        & \textbf{0.7}                       & \multicolumn{1}{c|}{\textbf{0.9}}                       & \textbf{Avg}                        \\ \hline
                                & QAT \cite{nag2021few}                              & 44.9                                  & 29.2                                   & \multicolumn{1}{c|}{11.2}                                  & 25.9                                  \\
\multirow{-2}{*}{1}             & QAT \cite{nag2021few} + \textbf{GAP}                & \cellcolor[HTML]{ECF4FF}\textbf{45.8} & \cellcolor[HTML]{ECF4FF}\textbf{30.0} & \multicolumn{1}{c|}{\cellcolor[HTML]{ECF4FF}\textbf{11.8}} & \cellcolor[HTML]{ECF4FF}\textbf{26.5} \\ \hline
                                & QAT \cite{nag2021few}                               & 51.8                                  & 32 .6                                 & \multicolumn{1}{c|}{11.9}                                  & 30.2                                  \\
\multirow{-2}{*}{5}             & QAT \cite{nag2021few} + \textbf{GAP}                & \cellcolor[HTML]{ECF4FF}\textbf{52.2} & \cellcolor[HTML]{ECF4FF}\textbf{32.9} & \multicolumn{1}{c|}{\cellcolor[HTML]{ECF4FF}\textbf{12.1}} & \cellcolor[HTML]{ECF4FF}\textbf{30.4} \\ \hline
\end{tabular}
\end{table}

\subsection{GAP in Zero-Shot Setting}
Similar to few-shot approaches, we can use GAP in zero-shot TAD setting. We consider a very recent zero-shot method STALE \cite{nag2022pclfm} and a 2-stage baseline (similar to \textit{Baseline-I} in \cite{nag2022pclfm}) on a challenging 50\% seen data split on Activitynet dataset. Since the 2-stage baseline includes proposal-generation as an intermediate step, we can apply GAP during training of the CLIP \cite{radford2021learning} pre-trained classifier in the second stage.  On the other hand, STALE is a single-stage approach hence we use GAP in the localization head during post-processing. From Table~\ref{tabsup:6} we observe a higher improvement using GAP for the baseline, indicating 2-stage approaches have localization-error propagation which can be partially resolved by using GAP. This reveals another advantage of our model design. 

\begin{table}[h]
\centering
\small
\caption{Effect of our GAP on zero-shot action detection methods on ActivityNet dataset in 50\% seen data setting. $\dagger$ indicates GAP is used during training.}
\label{tabsup:6}
\begin{tabular}{c|cccc}
\hline
                                  & \multicolumn{4}{c}{\textbf{mAP}}                                                                                                                                           \\ \cline{2-5} 
\multirow{-2}{*}{\textbf{Models}} & \textbf{0.5}                        & \textbf{0.75}                       & \multicolumn{1}{c|}{\textbf{0.95}}                       & \textbf{Avg}                        \\ \hline
Baseline                          & 28.0                                   & 16.4                                   & \multicolumn{1}{c|}{1.2 }                                  & 16.0                                  \\
Baseline$^{\dagger}$ + \textbf{GAP}             & \cellcolor[HTML]{A9FADF}\textbf{28.7} & \cellcolor[HTML]{A9FADF}\textbf{16.8} & \multicolumn{1}{c|}{\cellcolor[HTML]{A9FADF}\textbf{1.7}} & \cellcolor[HTML]{A9FADF}\textbf{16.5} \\ 
Baseline + \textbf{GAP}             & \cellcolor[HTML]{A9FADF}\textbf{28.2} & \cellcolor[HTML]{A9FADF}\textbf{16.6} & \multicolumn{1}{c|}{\cellcolor[HTML]{A9FADF}\textbf{1.3}} & \cellcolor[HTML]{A9FADF}\textbf{16.2} \\\hline
STALE \cite{nag2022pclfm}                            & 32.1                                   & 20.7                                  & \multicolumn{1}{c|}{5.9}                                  & 20.5                                  \\
STALE \cite{nag2022pclfm} + \textbf{GAP}                & \cellcolor[HTML]{A9FADF}\textbf{32.4} & \cellcolor[HTML]{A9FADF}\textbf{21.1} & \multicolumn{1}{c|}{\cellcolor[HTML]{A9FADF}\textbf{6.2}} & \cellcolor[HTML]{A9FADF}\textbf{20.8} \\ \hline
\end{tabular}
\end{table}

\section{Summary}
From the experiments we have performed so far, we draw several conclusions regarding the usefulness of our proposed GAP. Besides being effective for fully-supervised setting (Table~\ref{tab:main}), our GAP is also effective when there are \textbf{(i)} a large number of unlabeled training samples (refer to Table~\ref{tabsup:2}), \textbf{(ii)} unavailability of fine-grained annotation (refer to Table~\ref{tabsup:3}), \textbf{(iii)} only a few labeled samples (refer to Table~\ref{tabsup:4}), and \textbf{(iv)} no labeled samples (refer to Table~\ref{tabsup:6}). In all the above mentioned cases, the quantization error is more profound due to the design choice or problem setting which can be greatly reduced by using our GAP.
This verifies the generic usefulness of our method across a variety of settings.
% thus justifying its effectiveness. 

\section{Limitations}
\textcolor{black}{
Although GAP enjoys the flexibility of being a plug-and-play module it comes with a few limitations. While being model agnostic and simple, it does not have high gain when the temporal resolution is large (Table \ref{tab:size}), \eg, greater than 400 snippets. 
This is because, at high temporal resolutions there is no much quantization error due to more duration per instance,
and post-processing is hence less needed.
% cannot bring much gain and is less needed. 
Nonetheless, our GAP still gives  a gain of 0.1\% in avg mAP, which is a meaningful boost considering that the metric is very strict and the model performance is saturating. The snippet duration issue can only be solved if the snippet sampling procedure is automated based on the quantized error, which will be a good research direction for  future research. 
% Please note, in temporal action detection, having  an improvement of 0.1\% in avg mAP is significant considering this metric is highly strict and the model performance is saturating. It's often that a specific new method yield no more than 0.5\% gain over prior art (e.g., xxx) with non-trivial model specific complex addition. While our method is model agnostic and simple. This is because, at high temporal resolution there is no much quantization error due to more duration per instance, hence the post-processing cannot bring much gain and is less needed. 
}

\section{Conclusion}
For the first time we systematically investigated largely ignored yet significant problem of \textit{temporal quantization error} for temporal action detection in untrimmed videos. We not only revealed the genuine significance of this problem, but also
presented a novel \textit{Gaussian Aware Post-processing} (GAP) for more accurate model
inference.  Serving as a ready-to-use plug-in,
existing state-of-the-art TAD models can be seamlessly benefited without any algorithmic adaptation at a neglectable cost. 
% Apart from demonstrating empirically the importance of action boundary representation, w
We validated the performance benefits of GAP over a wide variety of contemporary
models on two challenging datasets. 
When model re-training is allowed,
more significant performance gain can be achieved 
without complex model redesign and change.

%%%%%%%%% REFERENCES
{\small
\bibliographystyle{ieee_fullname}
\bibliography{egbib}
}

\end{document}